\documentclass{article}
\usepackage{ijcai19}

\usepackage[utf8]{inputenc}
\usepackage{times}
\usepackage{soul}
\usepackage{url}
\usepackage{xcolor}
\usepackage[small]{caption}
\usepackage{graphicx}
\usepackage{amsmath}
\usepackage{booktabs}
\usepackage[english]{babel}

\usepackage[export]{adjustbox}

\usepackage{latexsym}
\usepackage{amsmath}
\usepackage{amssymb}
\usepackage{subcaption}
\usepackage{enumitem}

\usepackage[ruled,vlined,linesnumbered]{algorithm2e}

\usepackage[pdftitle={Out of Sight But Not Out of Mind: An Answer Set Programming Based Online Abduction Framework for Visual Sensemaking in Autonomous Driving},
pdfauthor={Jakob Suchan, Mehul Bhatt, and Srikrishna Varadarajan},
linkbordercolor=white,
bookmarksopen=false,
bookmarksnumbered=false]{hyperref}

\hypersetup{
    pdftitle={Out of Sight But Not Out of Mind: An Answer Set Programming Based Online Abduction Framework for Visual Sensemaking in Autonomous Driving}, 						
    pdfauthor={Jakob Suchan, Mehul Bhatt, and Srikrishna Varadarajan},     					
    pdfsubject={Artificial Intelligence; Computer Vision; Cognitive Vision; Autonomous Driving; Explainable AI}, 									
    pdfkeywords={Artificial Intelligence; Computer Vision; Cognitive Vision; Autonomous Driving; Explainable AI}, 						
    unicode=false,          									
    pdftoolbar=false,        									
    pdfmenubar=true,        								
    pdffitwindow=false,     									
    pdfstartview={FitH},    								
    pdfnewwindow=true,      								
    bookmarks=true,         								
    colorlinks=true,       									
    linkcolor=red!70!black,          							
    citecolor=blue!50!black,        							
    filecolor=blue,      									
    urlcolor=blue!80!black,          						
}

\let\oldnl\nl
\newcommand{\nonl}{\renewcommand{\nl}{\let\nl\oldnl}}

\newcommand{\codeMark}{{\raisebox{0pt}[0pt][0pt]{\raisebox{0.15ex}{\tiny\color{blue!80!black}$\blacktriangleright~~$}}}}
\newcommand{\codeTitle}[1]{\codeMark{{\tiny\sffamily\color{blue!70!black} \uppercase{#1}}}}

\newcommand{\predTh}[1]{{\operatorname{\mathsf{\color{blue!96!black}#1}}}}
\newcommand{\predThF}[1]{{\operatorname{\mathsf{#1}}}}

\newcommand{\timePoints}[0]{$\mathcal{\color{blue!96!black}T}$}

\newcommand{\objectSortNormal}[0]{$\mathcal{O}$}
\newcommand{\objectSort}[0]{$\mathcal{\color{blue!96!black}O}$}

\newcommand{\motionTrack}[0]{$\mathcal{\color{blue!96!black}MT}$}

\newcommand{\spatialPrimitivesNormal}[0]{$\mathcal{E}$}
\newcommand{\spatialPrimitives}[0]{$\mathcal{\color{blue!96!black}E}$}
\newcommand{\spatialRelations}[0]{$\mathcal{\color{blue!96!black}R}$}

\newcommand{\citet}[1]{\citeauthor{#1}~\shortcite{#1}}
\newcommand{\citep}{\cite}

\definecolor{YellowGreen}{RGB}{160,200,40}
\definecolor{mathcolor}{RGB}{7,72,110}

\title{Out of Sight But Not Out of Mind:\\[2pt]An Answer Set Programming Based Online Abduction Framework for\\Visual Sensemaking in Autonomous Driving}

\author{
Jakob Suchan$^{1,3}$\and 
Mehul Bhatt$^{2,3}$\And
Srikrishna Varadarajan$^{3}$
\affiliations
$^1$University of Bremen, Germany, $^2$\"{O}rebro University, Sweden\\[3pt]$^3$CoDesign Lab /  Cognitive Vision -- \url{www.codesign-lab.org}\\
}

\begin{document}

\maketitle

\begin{abstract}{
We  demonstrate the need and potential of systematically integrated \emph{vision} and \emph{semantics} solutions for visual sensemaking (in the backdrop of autonomous driving). 
A general method for \emph{online} visual sensemaking using answer set programming is systematically formalised and fully implemented. The method integrates state of the art in (deep learning based) visual computing, and is developed as a modular framework usable within hybrid architectures for perception \& control. We evaluate and demo with community established benchmarks {\small KITTIMOD} and {\small MOT}. As use-case, we focus on the significance of human-centred visual sensemaking ---e.g., semantic representation and explainability, question-answering, commonsense interpolation--- in safety-critical autonomous driving situations.
}
\end{abstract}

\section{\uppercase{Motivation}}
Autonomous driving research has received enormous academic \& industrial interest in recent years. This surge has coincided with (and been driven by) advances in \emph{deep learning} based computer vision research. Although {deep learning} based vision \& control has (arguably) been successful for self-driving vehicles, we posit that there is a clear need and tremendous potential for hybrid visual sensemaking solutions (integrating \emph{vision and semantics}) towards fulfilling essential legal and ethical responsibilities involving explainability, human-centred AI, and industrial standardisation (e.g, pertaining to representation, realisation of rules and norms).

\subsubsection{Autonomous Driving:$~~$``Standardisation \& Regulation''}
As the self-driving vehicle industry develops, it will be necessary ---e.g., similar to sectors such as medical computing, computer aided design--- to have an articulation and community consensus on aspects such as representation, interoperability, human-centred performance benchmarks, and data archival \& retrieval mechanisms.\footnote{{\sffamily\scriptsize Within autonomous driving, the need for standardisation and ethical regulation has most recently garnered interest internationally, e.g.,  with the Federal Ministry of Transport and Digital Infrastructure in Germany taking a lead in eliciting 20 key propositions (with legal implications) for the fulfilment of ethical commitments for automated and connected driving systems \citep{ethicalGermany2018}.}} In spite of major investments in self-driving vehicle research, issues related to human-centred'ness, human collaboration, and standardisation have been barely addressed, with  the current focus in driving research primarily being on two basic considerations: \emph{how fast to drive, and which way and how much to steer}. This is necessary, but inadequate if autonomous vehicles are to become commonplace and function with humans. Ethically driven standardisation and regulation will require addressing challenges in semantic visual interpretation, natural / multimodal human-machine interaction, high-level data analytics (e.g., for post hoc diagnostics, dispute settlement) etc. This will necessitate --amongst other things-- human-centred qualitative benchmarks and multifaceted hybrid AI solutions.

\begin{figure}[t]
\center
\includegraphics[width=0.95\columnwidth]{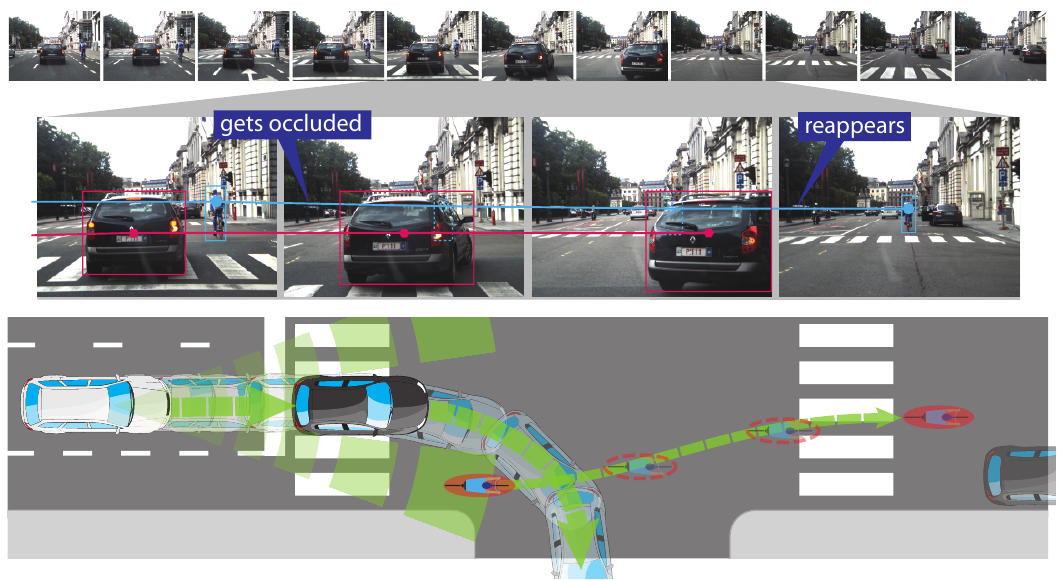}
\caption{{{\bf Out of sight but not out of mind}; the case of hidden entities: an occluded cyclist.}}
\label{fig:first-page-scanario}
\end{figure}

\begin{figure*}[t]
\center
\includegraphics[width = 0.8\textwidth]{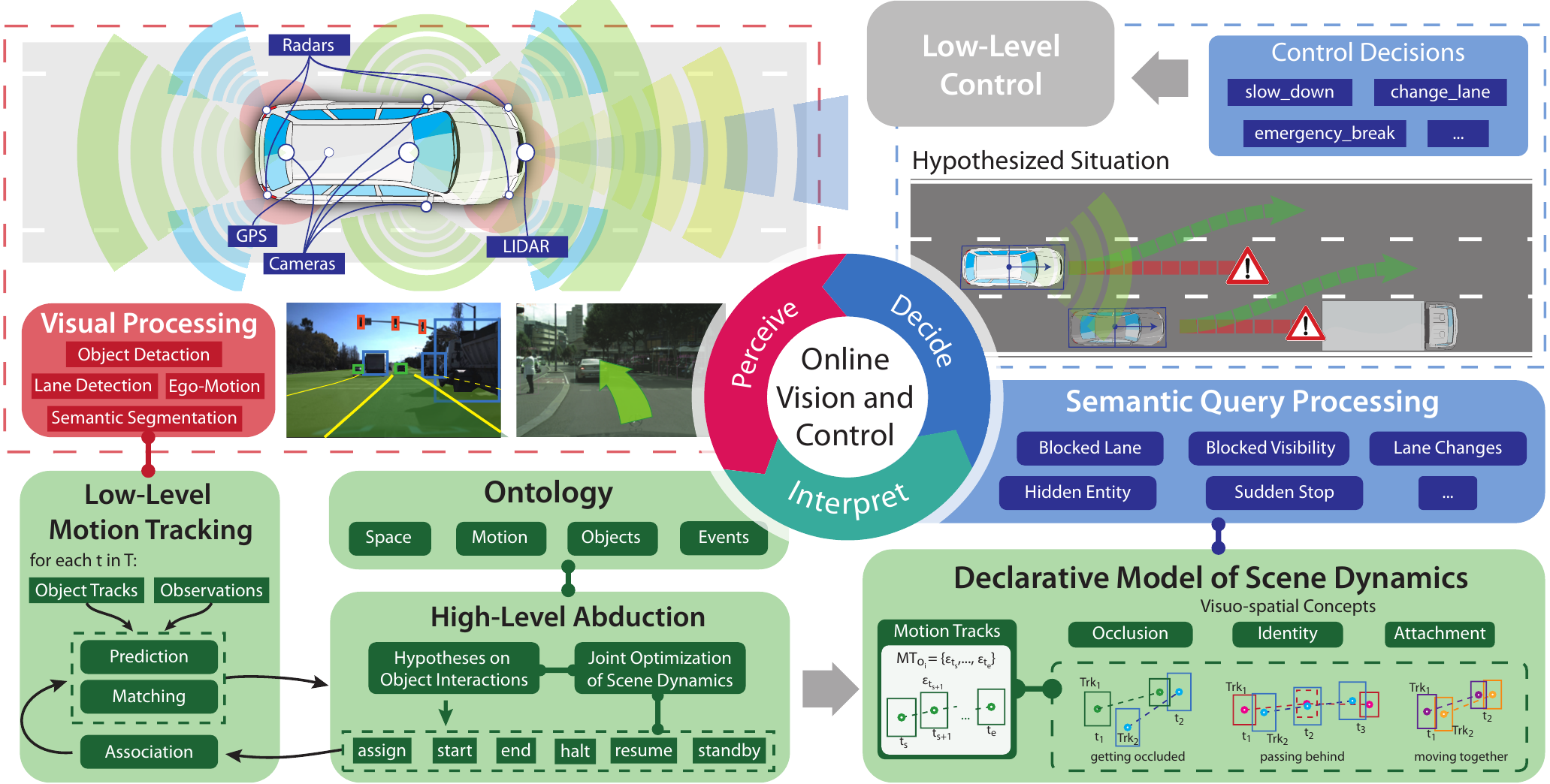}
\caption{{ A General Online Abduction Framework$~$ / $~$Conceptual Overview}}
\label{fig:system_overview}
\end{figure*}

\subsubsection{Visual Sensemaking Needs Both ``Vision \& Semantics''} 
We demonstrate the significance of semantically-driven methods rooted in knowledge representation and reasoning (KR) in addressing research questions pertaining to explainability and human-centred AI particularly from the viewpoint of sensemaking of dynamic visual imagery. Consider the \emph{occlusion scenario} in Fig. \ref{fig:first-page-scanario}:

\smallskip

\noindent{\footnotesize
Car ($c$) is {\color{blue!80!black}in-front}, and indicating to {\color{blue!80!black}turn-right}; {\color{blue!80!black}during} this time, person ($p$) is {\color{blue!80!black}on} a bicycle ($b$) and positioned {\color{blue!80!black}front-right} of $c$ and {\color{blue!80!black}moving-forward}. Car $c$ turns-right, during which the bicyclist $<p, b>$ is not {\color{blue!80!black}visible}. {\color{blue!80!black}Subsequently}, bicyclist $<p, b>$ {\color{blue!80!black}reappears}.}

\smallskip

\noindent The occlusion scenario indicates several challenges concerning aspects such as: {identity maintenance, making default assumptions, computing counterfactuals, projection, and interpolation of missing information} (e.g., what could be hypothesised about {\small\sffamily bicyclist $<p, b>$} when it is \emph{occluded}; how can this hypothesis enable in planning an immediate next step). Addressing such challenges ---be it realtime or post-hoc--- in view of human-centred AI concerns pertaining to ethics, explainability and regulation requires a systematic integration of {\small\textbf{Semantics and Vision}}, i.e., robust commonsense representation \& inference about spacetime dynamics on the one hand, and powerful low-level visual computing capabilities, e.g., pertaining to object detection and tracking on the other.

\smallskip

\noindent \textbf{Key Contributions}.\quad We develop a general and systematic declarative visual sensemaking method capable of online abduction: \emph{realtime, incremental, commonsense} semantic question-answering and belief maintenance over dynamic visuospatial imagery. Supported are {\small\bf(1--3)}:\quad {\small\bf(1)}. human-centric representations semantically rooted in  spatio-linguistic primitives as they occur in natural language \citep{Bhatt-Schultz-Freksa:2013,mani-james-motion}; \quad {\small\bf(2)}. driven by Answer Set Programming {\small (ASP)} \citep{Brewka:2011:ASP}, the ability to abductively compute commonsense interpretations and explanations in a range of (a)typical everyday driving situations, e.g., concerning safety-critical decision-making;\quad {\small\bf(3)}. online performance of the overall framework modularly integrating high-level commonsense reasoning and state of the art low-level visual computing for practical application in real world settings. We present the formal framework \& its implementation, and demo \& empirically evaluate with community established real-world datasets and benchmarks, namely: {\small KITTIMOD} \citep{Geiger2012CVPR} and {\small MOT} \citep{MOT16-Benchmark}.

\section{\uppercase{Visual Sensemaking}:\\\uppercase{A General Method Driven by ASP}}
Our proposed framework, in essence, jointly solves the problem of assignment of \emph{detections} to \emph{tracks} and explaining overall scene dynamics (e.g. {\footnotesize\sffamily appearance}, {\footnotesize\sffamily disappearance}) in terms of high-level \emph{events} within an online integrated low-level visual computing and high level abductive reasoning framework (Fig. \ref{fig:system_overview}). Rooted in answer set programming, the framework is general, modular, and designed for integration as a reasoning engine within (hybrid) architectures designed for real-time decision-making and control where visual perception is needed as one of the several components. In such large scale AI systems the declarative model of the scene dynamics resulting from the presented framework can be used for semantic {\small Q/A}, inference etc. to support decision-making.

\subsection{\uppercase{Space, Motion, Objects, Events}}
\label{sub_sec:space_motion}
Reasoning about dynamics is based on high-level representations of objects and their respective motion \& mutual interactions in spacetime. Ontological primitives for commonsense reasoning about spacetime ({\color{blue!80!black}$\Sigma_{st}$}) and dynamics ({\color{blue!80!black}$\Sigma_{dyn}$}) are:

\smallskip

\noindent $\bullet~~${\color{blue!80!black}$\Sigma_{st}$}: {\textbf{domain-objects}} \objectSort\ = $\{o_1, ... , o_n\}$ represent the visual elements in the scene, e.g., \emph{people}, \emph{cars}, \emph{cyclists}; elements in \objectSortNormal\ are geometrically interpreted as \textbf{spatial entities} \spatialPrimitives = $\{\varepsilon_{1}, ..., \varepsilon_{n}\}$;  spatial entities \spatialPrimitivesNormal\  may be regarded as \emph{points}, \emph{line-segments} or (axis-aligned) \emph{rectangles} based on their spatial properties (and a particular reasoning task at hand). The temporal dimension is represented by \textbf{time points} \timePoints $= \{t_{1}, ..., t_{n}\}$. {\color{blue}$\mathcal{MT}_{o_i}$} = ($\varepsilon_{t_{s}}, ..., \varepsilon_{t_{e}}$) represents the \textbf{motion track} of a single object $o_i$, where $t_s$ and $t_e $ denote the start and end time of the track and $\varepsilon_{t_s}$ to $\varepsilon_{t_e}$ denotes the spatial entity (\spatialPrimitivesNormal) ---e.g., the \emph{axis-aligned bounding box}---corresponding to the object $o_i$ at time points $t_s$ to $t_e $. The spatial configuration of the scene and changes within it are characterised based on the qualitative \textbf{spatio-temporal relationships} (\spatialRelations) between the domain objects. For the running and demo examples of this paper, positional relations on axis-aligned rectangles based on the rectangle algebra (RA) \cite{Balbiani1999} suffice; RA uses the relations of Interval Algebra (IA) \cite{Allen1983} {\footnotesize $\mathcal{R}_{\mathsf{IA}} \equiv {\color{mathcolor}\{}\mathsf{before},$ $\mathsf{after},$ $\mathsf{during},$ $\mathsf{contains},$ $\mathsf{starts},$ $\mathsf{started\_by},$ $\mathsf{finishes},$ $\mathsf{finished\_by},$ $\mathsf{overlaps},$ $\mathsf{overlapped\_by},$ $\mathsf{meets},$ $\mathsf{met\_by},$  $\mathsf{equal}{\color{mathcolor}\}}$} to relate two objects by the \emph{interval relations} projected along each dimension separately (e.g., horizontal and vertical dimensions). 

\smallskip

\noindent $\bullet~~${\color{blue!80!black}$\Sigma_{dyn}$}: The set of \textbf{fluents} ${\color{blue!96!black}\Phi} = \{\phi_1, ... , \phi_n\}$ and \textbf{events} ${\color{blue!96!black}\Theta} = \{\theta_1, ... , \theta_n\}$ respectively characterise the dynamic properties of the objects in the scene and high-level abducibles (Table \ref{tbl:events}). For reasoning about dynamics (with {\small$<${\color{blue!96!black}$\Phi$}}, {\small{\color{blue!96!black}$\Theta$}$>$}), we use a variant of event calculus as per \cite{Ma2014-ASP-based_Epistemic_Event_Calculus,Miller2013-Epistemic_Event_Calculus}; in particular, for examples of this paper, the functional event calculus fragment ($\Sigma_{dyn}$) of \citet{Ma2014-ASP-based_Epistemic_Event_Calculus} suffices: main axioms relevant here pertain to $\predTh{occurs-at}(\theta, t)$ denoting that an event occurred at time $t$ and $\predTh{holds-at}(\phi, v, t)$ denoting that $v$ holds for a fluent $\phi$ at time $t$.\footnote{\scriptsize\sffamily ASP encoding of the domain independent axioms of the Functional Event Calculus (FEC) used as per: \href{https://www.ucl.ac.uk/infostudies/efec/fec.lp}{https://www.ucl.ac.uk/infostudies/efec/fec.lp}}

\smallskip

\noindent $\bullet~~${\color{blue!80!black}$\Sigma$}:\quad {\small Let $\Sigma$ $~\equiv_{def}~$  $\Sigma_{dyn}$ {\small$<${\color{blue!96!black}$\Phi$}}, {\small{\color{blue!96!black}$\Theta$}$>$} $~\cup~$  $\Sigma_{st}$ {\small$<$\objectSort, \spatialPrimitives, \timePoints, \motionTrack, \spatialRelations$>$}}

{
\SetNlSty{texttt}{\color{blue}}{\quad}
\IncMargin{1.8em}
\begin{algorithm}[t]\small
\scriptsize

\KwData{
Visual imagery ({\color{blue!96!black}$\mathcal{V}$}), and \\background knowledge $\Sigma$ $~\equiv_{def}~$  $\Sigma_{dyn}$ $~\cup~$  $\Sigma_{st}$ 
	}
	  \smallskip

\KwResult{
Visual Explanations ({\color{blue!96!black}$\mathcal{EXP}$})\hfill {\color{blue!96!black}(also: Refer Fig \ref{fig:computational_steps})}
}

\smallskip


$\mathcal{MT} \leftarrow \varnothing$,  $\mathcal{H}^{events} \leftarrow \varnothing$

\smallskip

\For{$t \in T$}
{

\smallskip

	$\mathcal{VO}_{t} \leftarrow observe(\mathcal{V}_t)$
	
	\smallskip
	
	$\mathcal{P}_{t} \leftarrow \varnothing$, $\mathcal{ML}_{t}  \leftarrow \varnothing$
	
	\smallskip
	
	\For{$trk \in \mathcal{MT}_{t-1}$}
	{
		$p_{trk} \leftarrow kalman\_predict(trk)$
		
		$\mathcal{P}_{t} \leftarrow \mathcal{P}_{t} \cup p_{trk}$
		
		\smallskip
		
		\For{$obs \in \mathcal{VO}_{t}$}
		{
			$ml_{trk, obs} \leftarrow calc\_IoU(p_{trk}, obs)$
			
			$\mathcal{ML}_{t} \leftarrow \mathcal{ML}_{t} \cup ml_{trk, obs}$
		}
		
	}
	
	\smallskip

	Abduce$(<\mathcal{H}^{assign}_{t},~\mathcal{H}^{events}_{t}>)$, such that:\hfill{\color{blue!96!black}(Step 2)}
	
	{\nonl{{$~~~~\Sigma \wedge \mathcal{H}^{events} \wedge [\mathcal{H}^{assign}_{t} \wedge \mathcal{H}^{events}_{t}] \models \mathcal{VO}_{t} \wedge \mathcal{P}_{t} \wedge \mathcal{ML}_t  $}}}\\[4pt]
	$\mathcal{H}^{events} \leftarrow \mathcal{H}^{events} \cup \mathcal{H}^{events}_{t}$

	\smallskip
	
	$\mathcal{MT}_{t} \leftarrow update(\mathcal{MT}_{t-1}, \mathcal{VO}_{t}, \mathcal{H}_{assign})$

}

\Return{$~\mathcal{EXP} ~~ \leftarrow~~ <\mathcal{H}^{events}, \mathcal{MT}>$}

\caption{$~${\small\color{blue!96!black} ${Online\_Abduction}(\mathcal{V}, \Sigma)$}
\label{alg:VA}
}
\end{algorithm}
}

\begin{figure}[t]
\centering
\scriptsize
\includegraphics[width = \columnwidth]{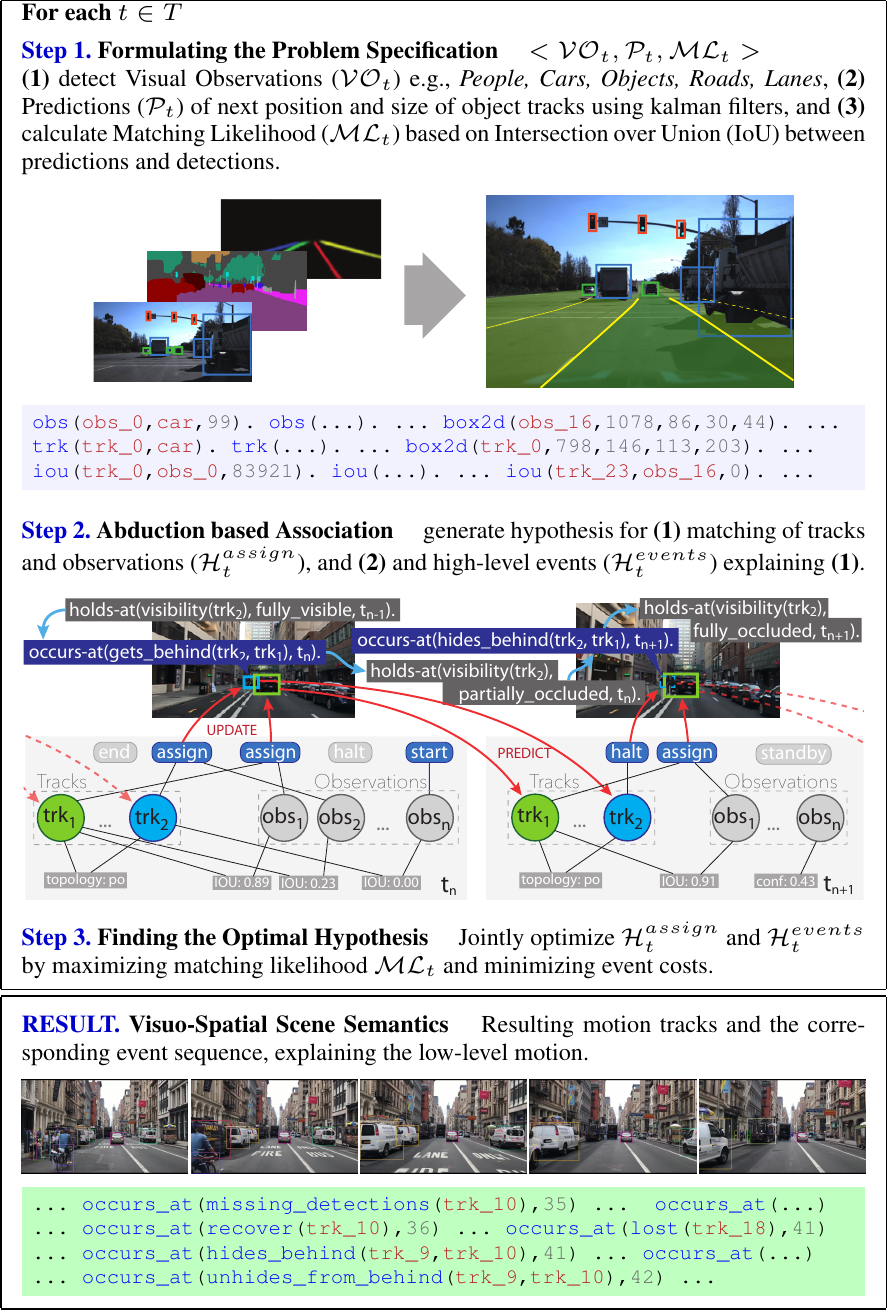}
\caption{{Computational Steps for Online Visual Abduction}}
\label{fig:computational_steps}
\end{figure}

\subsection{\uppercase{Tracking as Abduction}}\label{sec:trancking-as-abbduction}
Scene dynamics are tracked using a \emph{detect and track} approach: we tightly integrate low-level visual computing (for detecting scene elements) with high-level {\small ASP}-based abduction to solve the assignment of observations to object tracks in an \emph{incremental} manner. For each time point $t$ we generate a \emph{problem specification} consisting of the object tracks and visual observations and use {\small (ASP)} to abductively solve the corresponding assignment problem incorporating the ontological structure of the domain / data (abstracted with {$\Sigma$}). {\small\bf Steps 1--3} (Alg. \ref{alg:VA} \& Fig. \ref{fig:computational_steps}) are as follows:

\smallskip

\noindent {\small{\color{blue!80!black}\textbf{Step 1.}}\quad\small\uppercase{\textbf{Formulating the Problem Specification}}}\quad 
The {\small ASP} problem specification for each time point $t$ is given by the tuple {\small$<\mathcal{VO}_t, \mathcal{P}_t, \mathcal{ML}_t >$} and the sequence of events ($\mathcal{H}^{events}$) before time point $t$.

\smallskip

\noindent $\bullet~~${\small\textbf{Visual Observations}}\quad Scene elements derived directly from the visual input data are represented as spatial entities $\mathcal{E}$, i.e., $\mathcal{{\color{blue!96!black}VO}}_t$ = $\{\varepsilon_{obs_1}, ... , \varepsilon_{obs_n}\}$ is the set of observations at time $t$  (Fig. \ref{fig:computational_steps}). For the examples and empirical evaluation in this paper (Sec. \ref{sec:application-sec}) we focus on \emph{Obstacle / Object Detections} -- detecting {\small\sffamily cars, pedestrians, cyclists, traffic lights} etc using {\small YOLOv3} \cite{Redmon2018_YOLOv3}. Further we generate scene context using \emph{Semantic Segementation} -- segmenting the {\small\sffamily road, sidewalk, buildings, cars, people, trees}, etc. using {\small DeepLabv3+} \cite{deeplabv3plus2018}, and \emph{Lane Detection} -- estimating lane markings, to detect {\small\sffamily lanes} on the road, using {\small SCNN} \cite{Pan2018_scnn}. Type and confidence score for each observation is given by $type_{obs_i}$ and $conf_{obs_i}$.

\smallskip

\noindent $\bullet$\quad{\small\textbf{Movement Prediction}}\quad For each track $trk_i$ changes in \emph{position} and \emph{size} are predicted using kalman filters; this results in an estimate of the spatial entity $\varepsilon$ for the next time-point $t$ of each motion track  $\mathcal{P}_t$ = $\{\varepsilon_{trk_1}, ... , \varepsilon_{trk_n}\}$.

\smallskip

\noindent $\bullet$\quad{\small\textbf{Matching Likelihood}}\quad For each pair of tracks and observations $\varepsilon_{trk_i}$ and $\varepsilon_{obs_j}$, where $\varepsilon_{trk_i}\in\mathcal{P}_t$ and $\varepsilon_{obs_j}\in\mathcal{VO}_t$, we compute the likelihood $\mathcal{ML}_t = \{ml_{trk_1,obs_1}, ..., ml_{trk_i,obs_j}\}$ that $\varepsilon_{obs_j}$ belongs to $\varepsilon_{trk_i}$. The intersection over union (IoU) provides a measure for the amount of overlap between the \emph{spatial entities} $\varepsilon_{obs_j}$ and $\varepsilon_{trk_i}$.

\medskip

\noindent {\small{\color{blue!80!black}\textbf{Step 2.}}\quad\small\uppercase{\textbf{Abduction based Association }}}\quad 
Following perception as logical abduction most directly in the sense of \citet{Shanahan05}, we define the
task of abducing visual explanations as finding an association ($\mathcal{H}^{assign}_{t}$) of observed scene elements ($\mathcal{VO}_t$) to the motion tracks of objects ($\mathcal{MT}$) given by the predictions $\mathcal{P}_t$, together with a high-level explanation ($\mathcal{H}^{events}_{t}$), such that  $[\mathcal{H}^{assign}_{t} \wedge \mathcal{H}^{events}_{t}]$  is consistent with the background knowledge and the previously abduced event sequence $\mathcal{H}^{events}$, and entails the perceived scene given by $<\mathcal{VO}_t, \mathcal{P}_t, \mathcal{ML}_t >$:\\[3pt]\codeMark{\centering$~~~\Sigma \wedge \mathcal{H}^{events} \wedge [\mathcal{H}^{assign}_{t} \wedge \mathcal{H}^{events}_{t}] \models \mathcal{VO}_{t} \wedge \mathcal{P}_{t} \wedge \mathcal{ML}_t $}\\[2pt]where $\mathcal{H}^{assign}_{t}$ consists of the assignment of detections to object tracks, and $\mathcal{H}^{events}_{t}$ consists of the high-level \emph{events} {\small{$\Theta$}} explaining the assignments.

\smallskip

\noindent $\bullet~~${\small\textbf{Associating Objects and Observations}}\quad Finding the best match between observations ($\mathcal{VO}_t$) and object tracks ($\mathcal{P}_t$) is done by generating all possible assignments and then maximising a matching likelihood $ml_{trk_i,obs_j}$ between pairs of spatial entities for matched observations  $\varepsilon_{obs_j}$ and predicted track region $\varepsilon_{trk_i}$ (See Step 3). Towards this we use \emph{choice rules} \cite{Gebser2014-Clingo} (i.e., one of the heads of the rule has to be in the stable model) for $\varepsilon_{obs_j}$ and $\varepsilon_{trk_i}$, 
generating all possible assignments in terms of assignment actions:  \emph{assign, start, end, halt, resume, ignore\_det, ignore\_trk}.

\smallskip

\noindent\codeTitle{Matching tracks and detections}

\noindent\includegraphics[width = \columnwidth]{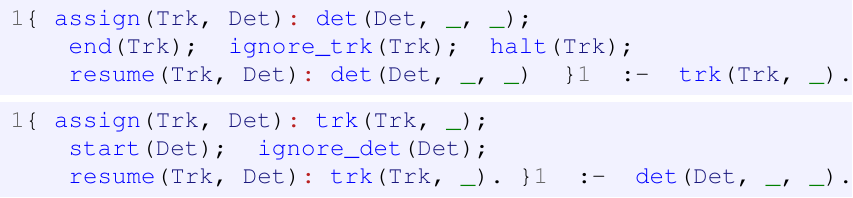}


\smallskip

\noindent For each assignment action we define integrity constraints\footnote{\scriptsize\sffamily Integrity constraints restrict the set of answers by eliminating stable models where the body is satisfied.} that restrict the set of answers generated by the choice rules, e.g., the following constraints are applied to assigning an observation $\varepsilon_{obs_j}$ to a track $trk_i$, applying thresholds on the $IoU_{trk_i,obs_j}$  and the confidence of the observation $conf_{obs_j}$, further we define that the type of the observation has to match the type of the track it is assigned to:

\smallskip

\noindent \codeTitle{Integrity constraints on matching}

\noindent \includegraphics[width = \columnwidth]{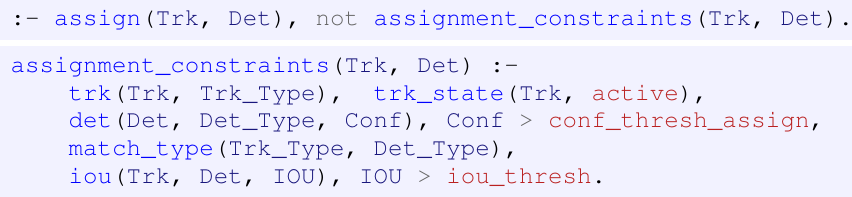}


\smallskip

\noindent $\bullet~~${\small\textbf{Abducible High-Level Events}}\quad  
For the length of this paper, we restrict to high-level visuo-spatial abducibles pertaining to \emph{object persistence} and \emph{visibility} (Table \ref{tbl:events}): {\small\bf(1)}. \emph{Occlusion}: Objects can disappear or reappear as result of occlusion with other objects; {\small\bf(2)}. \emph{Entering / Leaving the Scene}: Objects can enter or leave the scene at the borders of the field of view; {\small\bf(3)}.~\emph{Noise and Missing Observation}: (Missing-)\-observations can be the result of faulty detections.

\smallskip

Lets take the case of \emph{occlusion}:  functional fluent {\small\sffamily visibility} could be denoted $fully\_visible$, $partially\_occluded$ or $fully\_occluded$:

\smallskip

\noindent \codeTitle{Visibility - Fluent and Possible Values}

\noindent \includegraphics[width = \columnwidth]{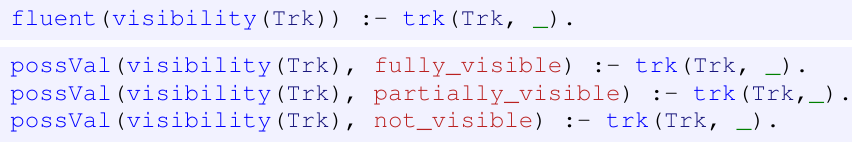}


\smallskip

\noindent We define the event $hides\_behind/2$, stating that an object hides behind another object by defining the conditions that have to hold for the event to possibly occur, and the effects the occurrence of the event has on the properties of the objects, i.e., the value of the visibility fluent changes to $fully\_occluded$.

{
\setlength{\textfloatsep}{3pt}
\begin{table}
\begin{center}
\scriptsize
\begin{tabular}{|l|p{1.7in}|}
\hline
\textbf{EVENTS} & \textbf{Description} \\\hline

$\predThF{enters\_fov(Trk)}$ & Track $\mathsf{Trk}$ enters the field of view.\\[2pt]

$\predThF{leaves\_fov(Trk)}$ & Track  $\mathsf{Trk}$ leaves the field of view.\\[2pt]

$\predThF{hides\_behind(Trk_1, Trk_2)}$ & Track $\mathsf{Trk_1}$ hides behind track $\mathsf{Trk_2}$.\\[2pt]

$\predThF{unhides\_from\_behind(Trk_1, Trk_2)}$ & Track $\mathsf{Trk_1}$ unhides from behind track $\mathsf{Trk_2}$.\\[2pt]

$\predThF{missing\_detections(Trk)}$ & Missing detections for track $\mathsf{Trk}$.\\[2pt]

\hline
\end{tabular}
\begin{tabular}{|p{0.98in}|p{0.75in}|p{1.23in}|}
\hline
\textbf{FLUENTS} & \textbf{Values} & \textbf{Description} \\\hline

$\predThF{in\_fov(Trk)}$ & \{true;false\} & Track  $\mathsf{Trk}$ is in the field of view.\\[2pt]

$\predThF{hidden\_by(Trk1, Trk2)}$ & \{true;false\} & Track  $\mathsf{Trk1}$ is hidden by $\mathsf{Trk2}$.\\[2pt]

$\predThF{visibility(Trk)}$ & \{fully\_visible; partially\_occluded; fully\_occluded\} & Visibility state of track  $\mathsf{Trk}$.\\\hline

\end{tabular}
\caption{{{\bf Abducibles}; Events and Fluents Explaining (Dis)Appearance}}
\label{tbl:events}
\end{center}
\end{table}%
}

%
%
%

\smallskip

\noindent \codeTitle{Occlusion - Event, Effects and (Spatial) Constraints}

\noindent \includegraphics[width = \columnwidth]{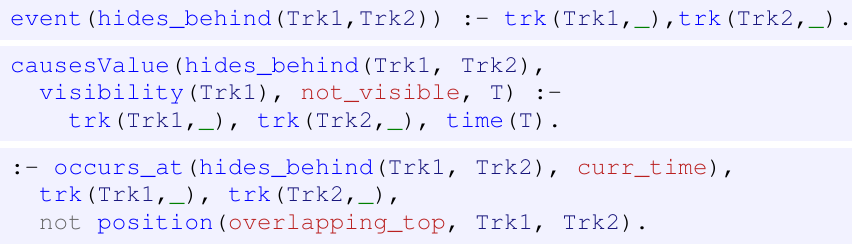}

%

\noindent For abducing the occurrence of an event we use choice rules that connect the event with assignment actions, e.g., a track getting halted may be explained by the event that the track hides behind another track.

\smallskip

\noindent \codeTitle{Generating Hypotheses on Events}

\noindent\includegraphics[width = \columnwidth]{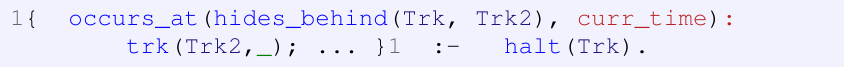}


\smallskip

\noindent {\small{\color{blue!80!black}\textbf{Step 3.}}\quad\small\uppercase{\textbf{Finding the Optimal Hypothesis}}}\quad  To ensure an \emph{optimal assignment}, we use {\small ASP} based optimization to maximize the matching likelihood between matched pairs of tracks and detections. Towards this, we first define the matching likelihood based on the Intersection over Union (IoU) between the observations and the predicted boxes for each track as described in \cite{Bewley2016_sort}:

\smallskip

\noindent \codeTitle{Assignment Likelihood}

\noindent \includegraphics[width = \columnwidth]{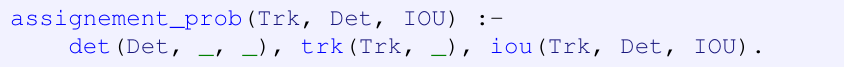}


\smallskip

We then maximize the matching likelihood for all assignments, using the build in \emph{maximize} statement:

\smallskip

\noindent \codeTitle{Maximizing Assignment Likelihood}

\noindent \includegraphics[width = \columnwidth]{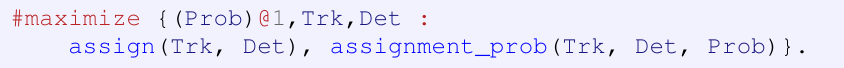}


\smallskip

\noindent To find the best set of hypotheses with respect to the observations, we \emph{minimize} the occurrence of certain events and association actions, e.g., the following optimization statements minimize starting and ending tracks; the resulting assignment is then used to update the motion tracks accordingly.

\smallskip

\noindent \codeTitle{Optimize Event and Association Costs}

\noindent \includegraphics[width = \columnwidth]{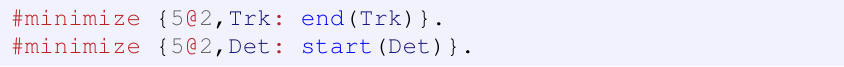}


\smallskip

\noindent It is important here to note that: {\small\textbf{(1)}}. by jointly abducing the object dynamics and high-level events we can impose constraints on the assignment of detections to tracks, i.e., an assignment is only possible if we can find an explanation supporting the assignment; and {\small\textbf{(2)}}. the likelihood that an event occurs guides the assignments of observations to tracks. Instead of independently tracking objects and interpreting the interactions, this yields to event sequences that are consistent with the abduced object tracks, and noise in the observations is reduced (See evaluation in Sec. \ref{sec:application-sec}).

{
\begin{table}[t]
\begin{center}
\tiny
\begin{tabular}{|l|p{1.3cm}|p{3.6cm}|}
\hline
\textbf{Situation} & \textbf{Objects} & \textbf{Description} \\\hline

OVERTAKING & \emph{vehicle, vehicle} & vehicle is overtaking another vehicle\\[2pt]

HIDDEN\_ENTITY & \emph{entity, object} & traffic participant hidden by obstacle\\[2pt]

REDUCED\_VISIBILITY & \emph{object} & visibility reduced by object in front.\\[2pt]

SUDDEN\_STOP & \emph{vehical} & vehicle in front stopping suddenly\\[2pt]

BLOCKED\_LANE & \emph{lane, object} & lane of the road is blocked by some object.\\[2pt]

EXITING\_VEHICLE & \emph{person, vehicle} & person is exiting a parked vehicle.\\\hline
\end{tabular}
\caption{{Safety-Critical Situations}}
\label{tbl:safety-critical-situations}
\end{center}
\end{table}
}

\section{\uppercase{Application \& Evaluation}}\label{sec:application-sec}
We demonstrate applicability towards identifying and interpreting \emph{safety-critical situations} (e.g., Table \ref{tbl:safety-critical-situations}); these encompass those scenarios where interpretation of spacetime dynamics, driving behaviour, environmental characteristics is necessary to anticipate and avoid potential dangers. 

\smallskip

\noindent \textbf{{{Reasoning about Hidden Entities}}}\quad
 Consider the situation of Fig. \ref{fig:example_occlusion}: a {\sffamily\small car} gets occluded by another car turning left and reappears \emph{in front of} the autonomous vehicle. Using online abduction for abducing high-level interactions of scene objects we can hypothesize that the {\sffamily\small car} got \emph{occluded} and anticipate its reappearance based on the perceived scene dynamics. The following shows data and abduced events. 

\smallskip

\noindent \includegraphics[width = \columnwidth]{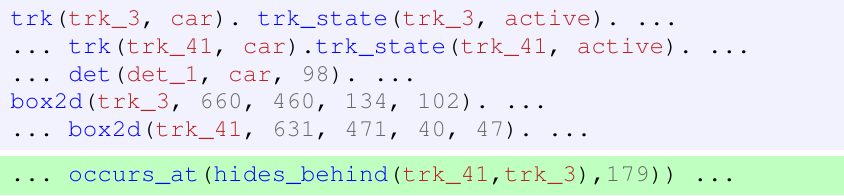}


\smallskip

\noindent We define a rule stating that a \emph{hidden} {\sffamily\small object} may \emph{unhide} from behind the object it is hidden by and anticipate the time point $t$ based on the object \emph{movement} as follows: 

\smallskip

\noindent \includegraphics[width = \columnwidth]{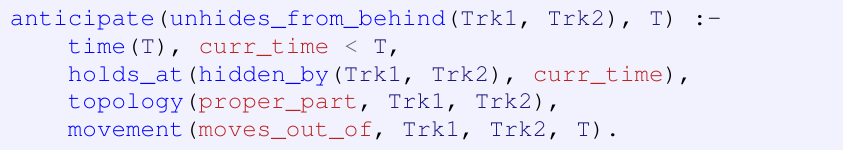}


\smallskip

\noindent We then interpolate the objects position at time point $t$ to predict where the object may \emph{reappear}.

\smallskip

\noindent \includegraphics[width = \columnwidth]{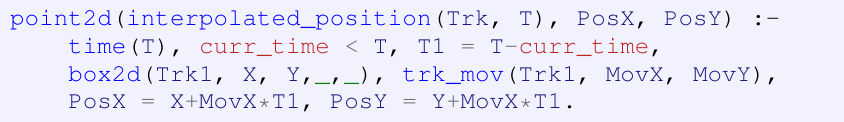}


\smallskip

\noindent For the occluded {\sffamily\small car} in our example we get the following prediction for time $t$ and position $x,y$:

\smallskip

\noindent \includegraphics[width = \columnwidth]{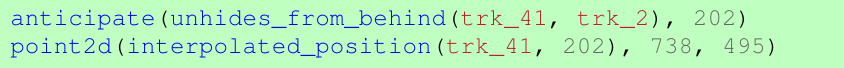}


\smallskip

\noindent Based on this prediction we can then define a rule that gives a warning if a hidden entity may reappear in front of the vehicle, which could be used by the control mechanism, e.g., to adapt driving and slow down in order to keep safe distance:

\smallskip

\noindent \includegraphics[width = \columnwidth]{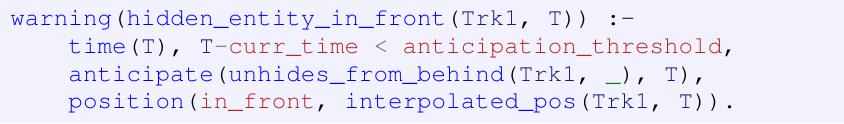}


{
\begin{figure}
\center

\includegraphics[width = \columnwidth]{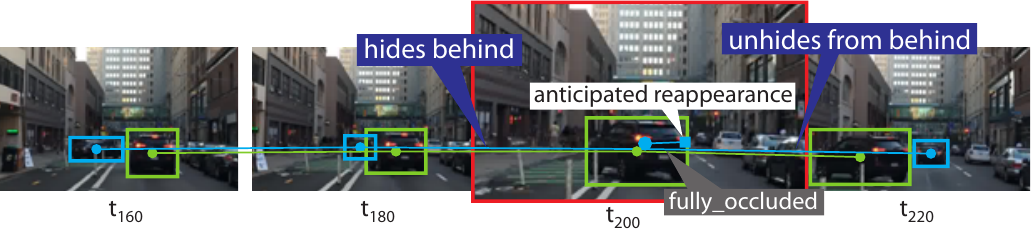}
\caption{{Abducing Occlusion to Anticipate Reappearance}}
\label{fig:example_occlusion}
\end{figure}
}

{
\begin{table*}[t]
\begin{center}
\scriptsize

\begin{tabular}{|p{3.5cm}|p{2.2cm}|c|c|c|c|c|c|c|c|}
\hline
\textbf{SEQUENCE} & Tracking & \textbf{MOTA} & \textbf{MOTP} & ML & MT & FP & FN & ID sw.  & Frag. \\\hline 

\hline
{\sffamily \scriptsize KITTI tracking } -- \emph{Cars}&
\scriptsize{without Abduction}&
45.72 \% &
76.89 \% &
19.14 \% &
23.04 \% &
785 &
11182 &
1097 &
1440 \\

\tiny{(8008 frames, 636 targets)} &
\scriptsize{with Abduction}& 
{\bf 50.5 \%} & 
74.76 \% & 
20.21 \% &
23.23 \% &
1311 &
10439 &
165 &
490 \\\hline

\hline
{\sffamily \scriptsize KITTI tracking } -- \emph{Pedestrians}&
\scriptsize{without Abduction}&
28.71 \% &
71.43 \% &
26.94 \% &
9.58 \% &
1261 &
6119 &
539 &
833 \\

\tiny{(8008 frames, 167 targets)} &
\scriptsize{with Abduction}& 
{\bf 32.57 \%} & 
70.68 \% & 
22.15 \% &
14.37 \% &
1899 &
5477 &
115 &
444 \\\hline

\hline
{\sffamily \scriptsize MOT 2017}&
\scriptsize{without Abduction}&
41.4 \% &
88.0 \% &
35.53 \% &
16.48 \% &
4877 &
60164 &
779 &
741 \\

\tiny{(5316 frames, 546 targets)} &
\scriptsize{with Abduction}& 
{\bf 46.2 \%} & 
87.9 \% & 
31,32 \% &
20.7 \% &
5195 &
54421 &
800 &
904 \\\hline

\end{tabular}

\end{center}
\caption{{ \textbf{ Evaluation of Tracking Performance}; accuracy (MOTA), precision (MOTP), mostly tracked (MT) and mostly lost (ML) tracks, false positives (FP), false negatives (FN), identity switches (ID Sw.), and fragmentation (Frag.). 
}}
\label{tbl:MOT_result}
\end{table*}
}

\medskip

\normalsize
\noindent \textbf{{Empirical Evaluation}}\quad  For online sensemaking, evaluation focusses on accuracy of abduced motion tracks, real-time performance, and the tradeoff between performance and accuracy. Our evaluation uses the \textbf{{\small KITTI}} \emph{object tracking dataset}  \cite{Geiger2012CVPR}, which is a community established benchmark dataset for autonomous cars: it consists of $21$ training and $29$ test scenes, and provides accurate track annotations for $8$ object classes (e.g., {\footnotesize\sffamily  car, pedestrian, van, cyclist}). We also evaluate tracking results using the more general cross-domain \textbf{\small Multi-Object Tracking} (MOT) dataset \cite{MOT16-Benchmark} established as part of the \emph{MOT Challenge}; it consists of $7$ training and $7$ test scenes which are highly unconstrained videos filmed with both static and moving cameras. We evaluate on the available groundtruth for training scenes of both KITTI using YOLOv3 detections, and MOT17 using the provided faster RCNN detections.

\smallskip

\noindent $\bullet~~${\small{\textbf{Evaluating Object Tracking}}}\quad  For evaluating \emph{accuracy} {\small(MOTA)} and \emph{precision} {\small(MOTP)} of abduced object tracks we follow the Clear{\small MOT} \cite{Bernardin2008} evaluation schema. {Results} (Table \ref{tbl:MOT_result}) show that jointly abducing high-level object interactions together with low-level scene dynamics increases the accuracy of the object tracks, i.e, we consistently observe an improvement of about $5\%$, from $45.72\%$ to $50.5\%$ for \emph{cars} and $28.71 \%$ to $32.57 \%$ for \emph{pedestrians} on KITTI, and from $41.4\%$ to $46.2\%$ on MOT.

\smallskip

\begin{figure}[t]
\begin{center}
\scriptsize
\begin{tabular}{|c|c|c|c|c|c|c|}
\hline
\textbf{DETECTOR} & Recall & MOTA & MOTP & $fps_{det}$ & $fps_{abd}$ \\\hline
\hline
{\sffamily \scriptsize YOLOv3}&
0.690 &
50.5 \%&
74.76 \% &
45 &
33.9 \\
{\sffamily \scriptsize SSD}&
0.599 &
30.63 \%&
77.4 \% &
8 &
46.7 \\
{\sffamily \scriptsize FRCNN}&
0.624 &
37.96 \%&
72.9 \% &
5 &
32.0 \\
\hline

\end{tabular}

\center
\includegraphics[width = 0.4\columnwidth,valign=c]{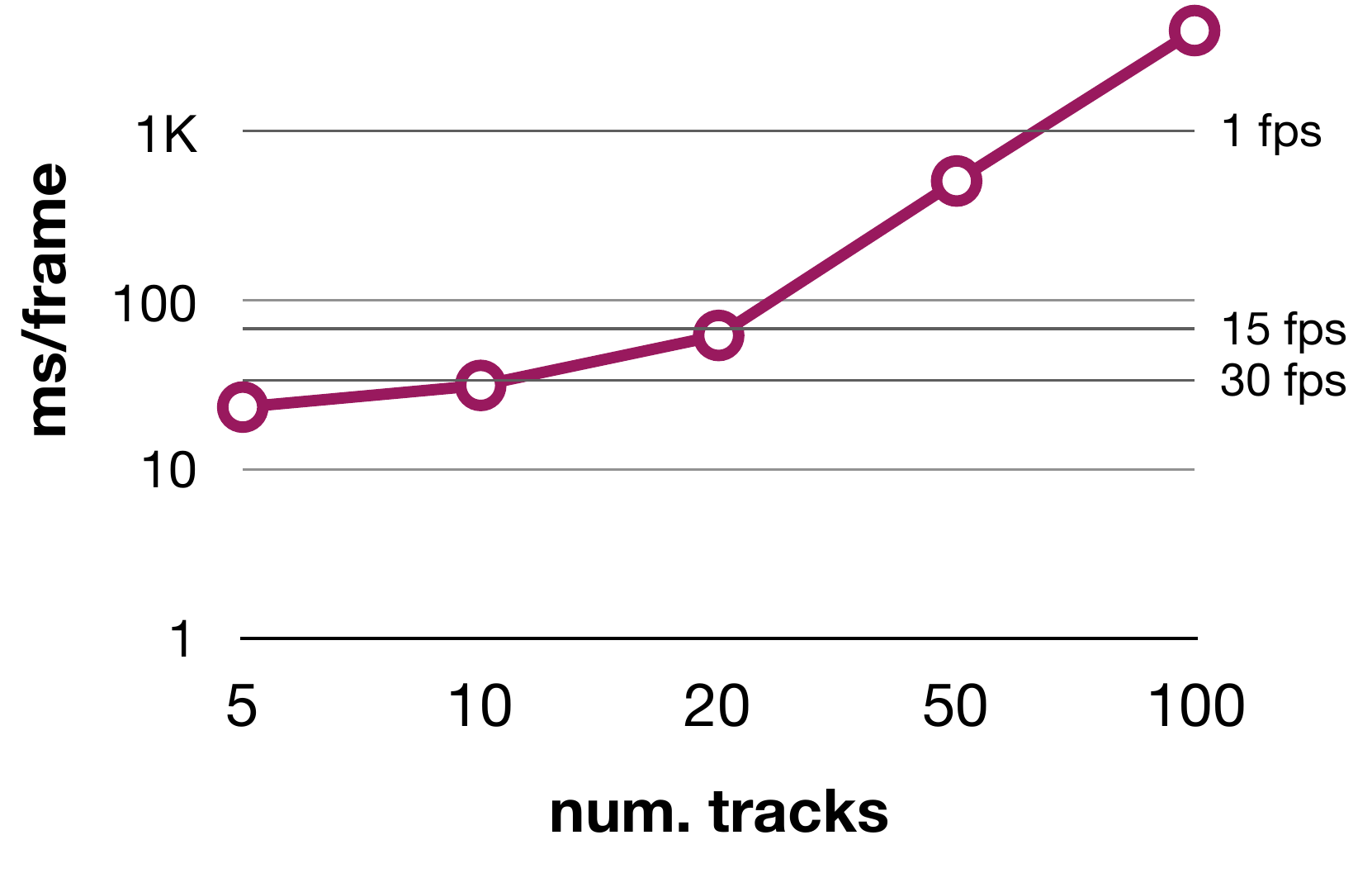}\quad
{
\tiny
\begin{tabular}{|c|r|r|c|c|c|c|}
\hline
\textbf{No. tracks} & $ms/frame$ & $fps$ \\\hline
\hline
{5}& 23.33 & 42.86 \\
{10}& 31.36 & 31.89  \\
{20}& 62.08 & 16.11 \\
{50}& 511.83 & 1.95 \\
{100}& 3996.38 & 0.25 \\
\hline
\end{tabular}
}

\end{center}
\caption{\textbf{Online Performance and Scalability}; performance for pretrained detectors on the \emph{'cars'} class of KITTI dataset, and processing time relative to the no. of tracks on synthetic dataset.}
\label{fig:real-time_result}
\end{figure}

\noindent $\bullet~~${\small{\textbf{Online Performance and Scalability}}}\quad Performance of online abduction is evaluated with respect to its real-time capabilities.\footnote{\scriptsize\sffamily Evaluation using a dedicated Intel Core i7-6850K 3.6GHz 6-Core Processor, 64GB RAM, and a NVIDIA Titan V GPU 12GB.}
{\small\bf (1).} We compare the time \& accuracy of online abduction for state of the art (real-time) detection methods: {\small YOLOv3}, 
{\small SSD} \cite{Liu16-ssd}, and {\small Faster RCNN} \cite{DBLP:conf/nips/RenHGS15} (Fig. \ref{fig:real-time_result}). 
{\small\bf (2).} We evaluate scalability of the ASP based abduction on a synthetic dataset with controlled number of tracks and \% of overlapping tracks per frame. Results  (Fig. \ref{fig:real-time_result}) show that online abduction can perform with above $30$ frames per second for scenes with up to $10$ highly overlapping object tracks, and more than 50 tracks with $1$fps (for the sake of testing, it is worth noting that even for $100$ objects per frame it only takes about an average of $4$ secs per frame). Importantly, for realistic scenes such as in the {\small KITTI} dataset, abduction runs realtime at $33.9$fps using {\small YOLOv3}, and $46.7$ using {\small SSD} with a lower accuracy but providing good precision.

\smallskip

\noindent \textbf{{Discussion of Empirical Results}}\quad  Results show that integrating high-level abduction and object tracking improves the resulting object tracks and reduce the noise in the visual observations. For the case of online visual sense-making, {\small ASP} based abduction provides the required performance: even though the complexity of {\small ASP} based abduction increases quickly, with large numbers of tracked objects the framework can track up to $20$ objects simultaneously with $30 fps$ and achieve real-time performance on the {\small KITTI} benchmark dataset. It is also important to note that the tracking approach in this paper is based on \emph{tracking by detection} using a naive measure, i.e, the IoU (Sec. \ref{sec:trancking-as-abbduction}; Step 1), to associate observations and tracks, and it is not using any visual information in the prediction or association step. Naturally, this results in a lower accuracy, in particular when used with noisy detections and when tracking fast moving objects in a benchmark dataset such as {\small KITTI}. That said, due to the modularity of the implemented framework, extensions with different methods for predicting motion (e.g., using particle filters or optical flow based prediction) are straightforward: i.e., improving tracking is not the aim of our research.

\section{\uppercase{Related Work}}
{\small ASP} is now widely used as an underlying knowledge representation language and robust methodology for non-monotonic reasoning \citep{Brewka:2011:ASP,Gebser2012-ASP}. With {\small ASP} as a foundation, and driven by semantics, commonsense and explainability \cite{DBLP:journals/cacm/DavisM15}, this research aims to bridge the gap between high-level formalisms for logical abduction and low-level visual processing by tightly integrating semantic abstractions of space-change with their underlying numerical representations. Within KR, the significance of  high-level (abductive) explanations in a range of contexts is well-established:  planning \& process recognition \cite{Kautz1991}, vision \& abduction \cite{Shanahan05}, probabilistic abduction \cite{Blythe11}, reasoning about spatio-temporal dynamics \cite{DBLP:journals/scc/BhattL08}, reasoning about continuous \emph{spacetime} change \cite{muller1998,Hazarika2002} etc. \citet{Dubba15} uses abductive reasoning in an inductive-abductive loop within inductive logic programming {\small (ILP)}. \citet{aditya2015visual} formalise general rules for image interpretation with {\small ASP}. Similarly motivated to this research is \citep{Suchan2018_visual_explanation}, which uses a two-step approach (with one huge \emph{problem specification}), first tracking and then explaining (and fixing) tracking errors; such an approach is not runtime / realtime capable. In computer vision research there has recently been an interest to synergise with cognitively motivated methods; in particular, e.g., for perceptual grounding \& inference \citep{compositional-grounding-jair2015} and combining video analysis with textual information for understanding events \& answering queries about video data \citep{DBLP:journals/ieeemm/TuMLCZ14}.

\section{\uppercase{Conclusion \& Outlook}}
We develop a novel abduction-driven \emph{online} (i.e., realtime, incremental) visual sensemaking framework: general, systematically formalised, modular and fully implemented. Integrating robust state-of-the-art methods in \emph{knowledge representation} and \emph{computer vision}, the framework has been evaluated and demonstrated with established benchmarks. We highlight application prospects of semantic vision for autonomous driving, a domain of emerging \& long-term significance. Specialised commonsense theories (e.g., about multi-sensory integration \& multi-agent belief merging, contextual knowledge) may be incorporated based on requirements. Our ongoing focus is to develop a novel dataset emphasising semantics and (commonsense) explainability; this is driven by mixed-methods research --AI, Psychology, HCI-- for the study of driving behaviour in low-speed, complex urban environments with unstructured traffic. Here, emphasis is on natural interactions (e.g., gestures, joint attention) amongst drivers, pedestrians, cyclists etc. Such interdisciplinary studies are needed to better appreciate the complexity and spectrum of varied human-centred challenges in autonomous driving, and demonstrate the significance of integrated vision \& semantics solutions in those contexts.

\smallskip

\vfill

{
\noindent \textbf{Acknowledgements}\\Partial funding by the German Research Foundation (DFG) via the CRC 1320 EASE -- Everyday Activity Science and Engineering'' (www.ease-crc.org), Project P3:  \emph{Spatial Reasoning in Everyday Activity} is acknowledged. 
}

\newpage

{
\small
\nocite{DBLP:conf/ijcai/SuchanB16}
\bibliographystyle{named}

\begin{thebibliography}{}

\bibitem[\protect\citeauthoryear{Aditya \bgroup \em et al.\egroup
  }{2015}]{aditya2015visual}
Somak Aditya, Yezhou Yang, Chitta Baral, Cornelia Fermuller, and Yiannis
  Aloimonos.
\newblock Visual commonsense for scene understanding using perception, semantic
  parsing and reasoning.
\newblock In {\em 2015 AAAI Spring Symposium Series}, 2015.

\bibitem[\protect\citeauthoryear{Allen}{1983}]{Allen1983}
James~F. Allen.
\newblock Maintaining knowledge about temporal intervals.
\newblock {\em Commun. ACM}, 26(11):832--843, 1983.


\bibitem[\protect\citeauthoryear{Balbiani \bgroup \em et al.\egroup
  }{1999}]{Balbiani1999}
Philippe Balbiani, Jean{-}Fran{\c{c}}ois Condotta, and Luis~Fari{\~{n}}as del
  Cerro.
\newblock A new tractable subclass of the rectangle algebra.
\newblock In Thomas Dean, editor, {\em {IJCAI} 1999, Sweden}, pages 442--447.
  Morgan Kaufmann, 1999.

\bibitem[\protect\citeauthoryear{Bernardin and
  Stiefelhagen}{2008}]{Bernardin2008}
Keni Bernardin and Rainer Stiefelhagen.
\newblock Evaluating multiple object tracking performance: The clear mot
  metrics.
\newblock {\em EURASIP Journal on Image and Video Processing}, 2008(1):246309,
  May 2008.

\bibitem[\protect\citeauthoryear{Bewley \bgroup \em et al.\egroup
  }{2016}]{Bewley2016_sort}
Alex Bewley, Zongyuan Ge, Lionel Ott, Fabio Ramos, and Ben Upcroft.
\newblock Simple online and realtime tracking.
\newblock In {\em 2016 IEEE International Conference on Image Processing
  (ICIP)}, pages 3464--3468, 2016.

\bibitem[\protect\citeauthoryear{Bhatt and
  Loke}{2008}]{DBLP:journals/scc/BhattL08}
Mehul Bhatt and Seng~W. Loke.
\newblock Modelling dynamic spatial systems in the situation calculus.
\newblock {\em Spatial Cognition {\&} Computation}, 8(1-2):86--130, 2008.

\bibitem[\protect\citeauthoryear{Bhatt \bgroup \em et al.\egroup
  }{2013}]{Bhatt-Schultz-Freksa:2013}
Mehul Bhatt, Carl Schultz, and Christian Freksa.
\newblock {The `Space' in Spatial Assistance Systems: Conception, Formalisation
  and Computation}.
\newblock In: {\em
  Representing space in cognition: Interrelations of behavior, language, and
  formal models. Series: Explorations in Language and Space}.
  978-0-19-967991-1, Oxford University Press, 2013.

\bibitem[\protect\citeauthoryear{Blythe \bgroup \em et al.\egroup
  }{2011}]{Blythe11}
James Blythe, Jerry~R. Hobbs, Pedro Domingos, Rohit~J. Kate, and Raymond~J.
  Mooney.
\newblock Implementing weighted abduction in markov logic.
\newblock In {\em Proc. of 9th Intl. Conference on
  Computational Semantics}, IWCS '11, USA, 2011. ACL.

\bibitem[\protect\citeauthoryear{{BMVI}}{2018}]{ethicalGermany2018}
{BMVI}.
\newblock Report by the ethics commission on automated and connected driving.,
  bmvi: Federal ministry of transport and digital infrastructure, germany,
  2018.

\bibitem[\protect\citeauthoryear{Brewka \bgroup \em et al.\egroup
  }{2011}]{Brewka:2011:ASP}
Gerhard Brewka, Thomas Eiter, and Miroslaw Truszczy\'{n}ski.
\newblock Answer set programming at a glance.
\newblock {\em Commun. ACM}, 54(12):92--103, December 2011.

\bibitem[\protect\citeauthoryear{Chen \bgroup \em et al.\egroup
  }{2018}]{deeplabv3plus2018}
Liang-Chieh Chen, Yukun Zhu, George Papandreou, Florian Schroff, and Hartwig
  Adam.
\newblock Encoder-decoder with atrous separable convolution for semantic image
  segmentation.
\newblock {\em arXiv:1802.02611}, 2018.

\bibitem[\protect\citeauthoryear{Davis and
  Marcus}{2015}]{DBLP:journals/cacm/DavisM15}
Ernest Davis and Gary Marcus.
\newblock Commonsense reasoning and commonsense knowledge in artificial
  intelligence.
\newblock {\em Commun. {ACM}}, 58(9):92--103, 2015.

\bibitem[\protect\citeauthoryear{Dubba \bgroup \em et al.\egroup
  }{2015}]{Dubba15}
Krishna Sandeep~Reddy Dubba, Anthony~G. Cohn, David~C. Hogg, Mehul Bhatt, and
  Frank Dylla.
\newblock Learning relational event models from video.
\newblock {\em J. Artif. Intell. Res. {(JAIR)}}, 53:41--90, 2015.

\bibitem[\protect\citeauthoryear{Gebser \bgroup \em et al.\egroup
  }{2012}]{Gebser2012-ASP}
Martin Gebser, Roland Kaminski, Benjamin Kaufmann, and Torsten Schaub.
\newblock {\em Answer Set Solving in Practice}.
\newblock Morgan \& Claypool Publishers, 2012.

\bibitem[\protect\citeauthoryear{Gebser \bgroup \em et al.\egroup
  }{2014}]{Gebser2014-Clingo}
Martin Gebser, Roland Kaminski, Benjamin Kaufmann, and Torsten Schaub.
\newblock Clingo = {ASP} + control: Preliminary report.
\newblock {\em CoRR}, abs/1405.3694, 2014.

\bibitem[\protect\citeauthoryear{Geiger \bgroup \em et al.\egroup
  }{2012}]{Geiger2012CVPR}
Andreas Geiger, Philip Lenz, and Raquel Urtasun.
\newblock Are we ready for autonomous driving? the kitti vision benchmark
  suite.
\newblock In {\em Conference on Computer Vision and Pattern Recognition
  (CVPR)}, 2012.

\bibitem[\protect\citeauthoryear{Hazarika and Cohn}{2002}]{Hazarika2002}
Shyamanta~M Hazarika and Anthony~G Cohn.
\newblock Abducing qualitative spatio-temporal histories from partial
  observations.
\newblock In {\em KR}, pages 14--25, 2002.

\bibitem[\protect\citeauthoryear{Kautz}{1991}]{Kautz1991}
Henry~A. Kautz.
\newblock Reasoning about plans.
\newblock chapter A Formal Theory of Plan Recognition and Its Implementation,
  pages 69--124. Morgan Kaufmann Publishers Inc., USA, 1991.

\bibitem[\protect\citeauthoryear{Liu \bgroup \em et al.\egroup
  }{2016}]{Liu16-ssd}
Wei Liu, Dragomir Anguelov, Dumitru Erhan, Christian Szegedy, Scott~E. Reed,
  Cheng{-}Yang Fu, and Alexander~C. Berg.
\newblock {SSD:} single shot multibox detector.
\newblock In {\em {ECCV} {(1)}}, volume 9905 of {\em LNCS}, pages 21--37. Springer, 2016.

\bibitem[\protect\citeauthoryear{Ma \bgroup \em et al.\egroup
  }{2014}]{Ma2014-ASP-based_Epistemic_Event_Calculus}
Jiefei Ma, Rob Miller, Leora Morgenstern, and Theodore Patkos.
\newblock An epistemic event calculus for asp-based reasoning about knowledge
  of the past, present and future.
\newblock In {\em {LPAR}: 19th Intl. Conf. on Logic for Programming, Artificial
  Intelligence and Reasoning}, volume~26 of {\em EPiC Series in Computing},
  pages 75--87. EasyChair, 2014.

\bibitem[\protect\citeauthoryear{Mani and
  Pustejovsky}{2012}]{mani-james-motion}
Inderjeet Mani and James Pustejovsky.
\newblock {\em Interpreting Motion - Grounded Representations for Spatial
  Language}, volume~5 of {\em Explorations in language and space}.
\newblock Oxford University Press, 2012.

\bibitem[\protect\citeauthoryear{Milan \bgroup \em et al.\egroup
  }{2016}]{MOT16-Benchmark}
Anton Milan, Laura Leal{-}Taix{\'{e}}, Ian~D. Reid, Stefan Roth, and Konrad
  Schindler.
\newblock {MOT16:} {A} benchmark for multi-object tracking.
\newblock {\em CoRR}, abs/1603.00831, 2016.

\bibitem[\protect\citeauthoryear{Miller \bgroup \em et al.\egroup
  }{2013}]{Miller2013-Epistemic_Event_Calculus}
Rob Miller, Leora Morgenstern, and Theodore Patkos.
\newblock Reasoning about knowledge and action in an epistemic event calculus.
\newblock In {\em {COMMONSENSE 2013}}, 2013.

\bibitem[\protect\citeauthoryear{Muller}{1998}]{muller1998}
Philippe Muller.
\newblock A qualitative theory of motion based on spatio-temporal primitives.
\newblock In Anthony G.~Cohn et. al., editor, {\em {KR 1998}, {I}taly}. Morgan
  Kaufmann, 1998.

\bibitem[\protect\citeauthoryear{Pan \bgroup \em et al.\egroup
  }{2018}]{Pan2018_scnn}
Xingang Pan, Jianping Shi, Ping Luo, Xiaogang Wang, and Xiaoou Tang.
\newblock Spatial as deep: Spatial {CNN} for traffic scene understanding.
\newblock In Sheila~A. McIlraith and Kilian~Q. Weinberger, editors, {\em {AAAI}
  2018}. {AAAI} Press, 2018.

\bibitem[\protect\citeauthoryear{Redmon and Farhadi}{2018}]{Redmon2018_YOLOv3}
Joseph Redmon and Ali Farhadi.
\newblock Yolov3: An incremental improvement.
\newblock {\em CoRR}, abs/1804.02767, 2018.

\bibitem[\protect\citeauthoryear{Ren \bgroup \em et al.\egroup
  }{2015}]{DBLP:conf/nips/RenHGS15}
Shaoqing Ren, Kaiming He, Ross~B. Girshick, and Jian Sun.
\newblock Faster {R-CNN:} towards real-time object detection with region
  proposal networks.
\newblock In {\em Annual Conference on Neural Information Processing Systems
  2015, Canada}, 2015.

\bibitem[\protect\citeauthoryear{Shanahan}{2005}]{Shanahan05}
Murray Shanahan.
\newblock Perception as abduction: Turning sensor data into meaningful
  representation.
\newblock {\em Cognitive Science}, 29(1):103--134, 2005.

\bibitem[\protect\citeauthoryear{Suchan and
  Bhatt}{2016}]{DBLP:conf/ijcai/SuchanB16}
Jakob Suchan and Mehul Bhatt.
\newblock Semantic question-answering with video and eye-tracking data: {AI}
  foundations for human visual perception driven cognitive film studies.
\newblock In S. Kambhampati, editor, {\em {IJCAI} 2016, New
  York, USA}, pages 2633--2639. {IJCAI/AAAI} Press, 2016.

\bibitem[\protect\citeauthoryear{Suchan \bgroup \em et al.\egroup
  }{2018}]{Suchan2018_visual_explanation}
Jakob Suchan, Mehul Bhatt, Przemyslaw~Andrzej Walega, and Carl P.~L. Schultz.
\newblock Visual explanation by high-level abduction: On answer-set programming
  driven reasoning about moving objects.
\newblock In: {\em {AAAI}
  2018}. {AAAI} Press, 2018.

\bibitem[\protect\citeauthoryear{Tu \bgroup \em et al.\egroup
  }{2014}]{DBLP:journals/ieeemm/TuMLCZ14}
Kewei Tu, Meng Meng, Mun~Wai Lee, Tae~Eun Choe, and Song~Chun Zhu.
\newblock Joint video and text parsing for understanding events and answering
  queries.
\newblock {\em {IEEE} MultiMedia}, 2014.

\bibitem[\protect\citeauthoryear{Yu \bgroup \em et al.\egroup
  }{2015}]{compositional-grounding-jair2015}
Haonan Yu, N.~Siddharth, Andrei Barbu, and Jeffrey~Mark Siskind.
\newblock {A Compositional Framework for Grounding Language Inference,
  Generation, and Acquisition in Video}.
\newblock {\em J. Artif. Intell. Res. {(JAIR)}}, 52:601--713, 2015.

\end{thebibliography}

}
\end{document}